\documentclass[10pt,twocolumn,letterpaper]{article}

\usepackage[pagenumbers]{cvpr} 

\usepackage{graphicx}
\usepackage{amsmath}
\usepackage{amssymb}
\usepackage{booktabs}
\usepackage{xcolor}
\usepackage[accsupp]{axessibility}  

\usepackage{bm}
\usepackage[pagebackref,breaklinks,colorlinks]{hyperref}

\usepackage[capitalize]{cleveref}
\crefname{section}{Sec.}{Secs.}
\Crefname{section}{Section}{Sections}
\Crefname{table}{Table}{Tables}
\crefname{table}{Tab.}{Tabs.}

\usepackage{wrapfig}
\usepackage{ifthen}
\newboolean{showNotes}
\setboolean{showNotes}{false}

\usepackage{xspace}

\newcommand{\Rahul}[1]{{\color{green}\textbf{Rahul}: #1}}

\newcommand{\Adrien}[1]{\textbf{\textcolor{orange}{Adrien: #1}}}

\newcommand{\norm}[1]{\left|\left| #1 \right|\right|}

\crefname{section}{Section}{Section}
\crefname{equation}{Equation}{Equation}
\crefname{figure}{Figure}{Figure}
\crefname{table}{Table}{Table}

\newcommand{\parahead}[1]{\noindent\textbf{#1}:\ }

\newenvironment{packed_itemize}
{\begin{itemize}
    \setlength{\itemsep}{1pt}
    \setlength{\parskip}{0pt}
    \setlength{\parsep}{0pt}
}{\end{itemize}}

\newcommand{\filluptopage}[1]{%
  \clearpage
  \loop\ifnum\value{page}<#1\relax
    \null\clearpage
  \repeat
  \loop\ifnum\value{page}=#1\relax
    \null\clearpage
  \repeat
}

\makeatletter
\def\blfootnote{\xdef\@thefnmark{}\@footnotetext}
\makeatother

\usepackage{xcolor}

\definecolor{cb-black}      {RGB}{  0,   0,   0}
\definecolor{cb-blue-green} {RGB}{  0,  073,  073}
\definecolor{cb-green-sea}  {RGB}{  0, 146, 146}
\definecolor{cb-rose}       {RGB}{255, 109, 182}
\definecolor{cb-salmon-pink}{RGB}{255, 182, 119}
\definecolor{cb-purple}     {RGB}{ 73,   0, 146}
\definecolor{cb-blue}       {RGB}{ 0, 109, 219}
\definecolor{cb-lilac}      {RGB}{182, 109, 255}
\definecolor{cb-blue-sky}   {RGB}{109, 182, 255}
\definecolor{cb-blue-light} {RGB}{182, 219, 255}
\definecolor{cb-burgundy}   {RGB}{146,   0,   0}
\definecolor{cb-brown}      {RGB}{146,  73,   0}
\definecolor{cb-clay}       {RGB}{219, 209,   0}
\definecolor{cb-green-lime} {RGB}{ 36, 255,  36}
\definecolor{cb-yellow}     {RGB}{255, 255, 109}

\usepackage{subcaption}

\def\cvprPaperID{9580} 
\def\confName{CVPR}
\def\confYear{2023}

\begin{document}

\newcommand{\ShortName}{Canonical Fields\xspace}
\newcommand{\MethodName}{CaFi-Net\xspace}
\title{\ShortName: Self-Supervised Learning of Pose-Canonicalized Neural Fields}

\author{
Rohith Agaram$^1$ \qquad Shaurya Dewan$^1$ \qquad Rahul Sajnani$^{2}$ \qquad Adrien Poulenard$^{3}$ \\ \qquad Madhava Krishna$^{1}$ \qquad Srinath Sridhar$^{2}$ \\ \vspace{1mm}
\text{\normalsize $^1$RRC, IIIT-Hyderabad \qquad $^2$Brown University \qquad $^3$Stanford University}\\
\href{https://ivl.cs.brown.edu/projects/canonicalfields/}{ivl.cs.brown.edu/projects/canonicalfields}
}
\twocolumn[{%
\renewcommand\twocolumn[1][]{#1}%
\maketitle
\begin{center}
    \vspace{-1.0cm}
    \centering
    \captionsetup{type=figure}
    \includegraphics[width=\textwidth]{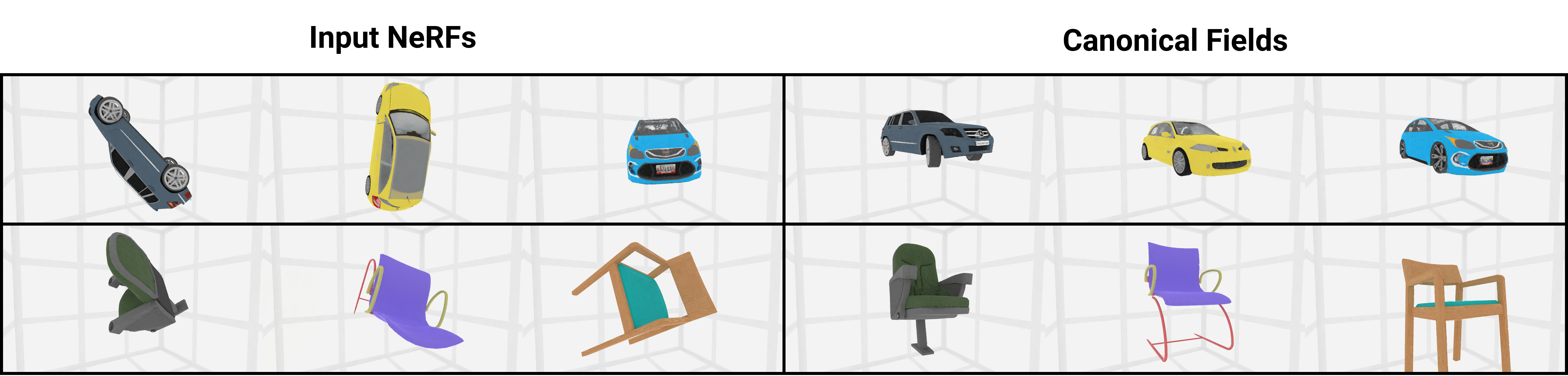}
    \vspace{-0.25in}
    \captionof{figure}{
    We present \underline{Ca}nonical \underline{Fi}eld Network (\textbf{\MethodName}), a self-supervised method for 3D position and orientation (pose) canonicalization of objects represented as neural fields.
    We specifically focus on neural radiance fields (NeRFs) fitted to raw RGB images of arbitrarily posed objects (left), and obtain a \textbf{canonical field} (fixed novel view shown on right) where all objects in a category are consistently positioned and oriented.
    \vspace{-0.25cm}
    }
    \label{fig:teaser}
\end{center}%
}]

\begin{abstract}
Coordinate-based implicit neural networks, or neural fields, have emerged as useful representations of shape and appearance in 3D computer vision.
Despite advances, however, it remains challenging to build neural fields for categories of objects without datasets like ShapeNet that provide ``canonicalized'' object instances that are consistently aligned for their 3D position and orientation (pose).
We present \underline{Ca}nonical \underline{Fi}eld Network (\textbf{\MethodName}), a self-supervised method to canonicalize the 3D pose of instances from an object category represented as neural fields, specifically neural radiance fields (NeRFs).
\MethodName directly learns from continuous and noisy radiance fields using a Siamese network architecture that is designed to extract equivariant field features for category-level canonicalization.
During inference, our method takes pre-trained neural radiance fields of novel object instances at arbitrary 3D pose and estimates a canonical field with consistent 3D pose across the entire category.
Extensive experiments on a new dataset of 1300 NeRF models across 13 object categories
show that our method matches or exceeds the performance of
3D point cloud-based methods.
%
\end{abstract}
\section{Introduction}
\label{sec:intro}
Neural fields~\cite{xie2022neuralfields}---coordinate-based neural networks that implicitly parameterize signals---have recently gained significant attention as representations of 3D shape~\cite{park2019deepsdf,Occupancy_Networks,chen2018implicit_decoder}, view-dependent appearance~\cite{mildenhall2020nerf,sitzmann2019_srn}, and motion~\cite{OccupancyFlow}.
In particular, neural radiance fields (NeRFs)~\cite{mildenhall2020nerf}, have been successfully used in problems such as novel view synthesis~\cite{barron2021mip,barron2022mip,zhang2020nerf++}, scene geometry extraction~\cite{yariv2021volsdf,wang2021neus}, capturing dynamic scenes~\cite{park2021nerfies,park2021hypernerf,tretschk2021non,peng2021animatable,li2022neural}, 3D semantic segmentation~\cite{zhi2021semanticnerf,vora2021nesf}, and robotics~\cite{adamkiewicz2022vision,ichnowski2021dexnerf,li20213d}.

Despite the progress, it remains challenging to build neural fields that represent an entire category of objects.
Previous methods sidestep the problem by overfitting on a single instance~\cite{mildenhall2020nerf}, or learning~\cite{park2019deepsdf,yu2021pixelnerf,Occupancy_Networks} on datasets like ShapeNet~\cite{shapenet2015} that contain objects that are manually \textbf{canonicalized} -- oriented consistently for 3D position and orientation (3D pose) across a category.
This strong supervision makes it easier to learn over categories,
but limits their application to data that contain these labels.
Recent work has proposed methods for self-supervised learning of 3D pose canonicalization~\cite{spezialetti2020learning,sun2020canonical,simeonov2022neural}, however, these operate on 3D point clouds, meshes, or voxels -- but not neural fields.

In this paper, we present \underline{Ca}nonical \underline{Fi}eld Network (\textbf{\MethodName}), a self-supervised method for category-level canonicalization of the 3D position and orientation of objects represented as \textbf{neural fields}, specifically neural radiance fields.
Canonicalizing neural fields is challenging because,
unlike 3D point clouds or meshes, neural fields are \textbf{continuous}, noisy, and hard to manipulate since they are parameterized as the weights of a neural network~\cite{yang2021geometry}.
To address these challenges, we first extend the notion of \textbf{equivariance} to continuous vector fields and show how networks for processing 3D point clouds~\cite{thomas2018tensor} can be extended to operate directly on neural radiance fields.
We design \MethodName as a Siamese network
that contains layers to extract equivariant features directly on vector fields.
These field features are used to learn a canonical frame that is consistent across instances in the category.
During inference, our method takes as input neural radiance fields of object instances from a category at arbitrary pose and estimates a \textbf{canonical field} that is consistent across the category.
To handle noise in radiance fields from NeRF, our method incorporates density-based feature weighting and foreground-background clustering.

Our approach learns canonicalization without any supervision labels on a new dataset of 1300 pre-trained NeRF models of 13 common ShapeNet categories in arbitrary 3D pose (see \cref{fig:teaser}).
We introduce several self-supervision loss functions that encourage the estimation of a consistent canonical pose.
%
In addition, we present extensive quantitative comparisons with baselines and other methods on standardized canonicalization metrics~\cite{sajnani2022condor} over 13 object categories.
In particular, we show that our approach matches or exceeds the performance of
3D point cloud-based methods.
This enables the new capability of directly operating on neural fields rather than converting them to point clouds for canonicalization.
To sum up, we contribute:
\begin{packed_itemize}
\item \underline{Ca}nonical \underline{Fi}eld Network (\textbf{\MethodName}), the first method for self-supervised canonicalization of the 3D position and orientation (pose) of objects represented as neural radiance fields.
\item A Siamese neural network architecture with equivariant feature extraction layers that are designed to directly operate on continuous and noisy radiance fields from NeRF.
\item A public dataset of 1300 NeRF models from 13 ShapeNet categories including posed images, and weights for evaluating canonicalization performance.
%
\end{packed_itemize}

\section{Related Work}
\label{sec:relwork}
We focus our review of related work on neural fields, supervised canonicalization, equivariant neural network architectures, and self-supervised canonicalization.

\parahead{Neural Fields}
Neural fields are emerging as useful representations for solving problems such as novel view synthesis~\cite{mildenhall2020nerf,barron2021mip,barron2022mip,zhang2020nerf++,jain2021putting}, shape encoding and reconstruction~\cite{park2019deepsdf,Occupancy_Networks}, dynamic reconstruction~\cite{park2021nerfies,park2021hypernerf,li2022neural}, appearance modeling~\cite{mildenhall2022nerf,rudnev2022nerfosr,srinivasan2021nerv}, and human motion modeling~\cite{tiwari22posendf,li2022tava,su2021nerf,corona2022lisa}.
Generalization of neural fields to object categories remains difficult, but some methods have used pre-canonicalized datasets to circumvent this problem~\cite{yu2021pixelnerf}.
Neural fields are also used outside of computer vision, for instance in reconstructing proteins~\cite{zhong2021cryodrgn2}, physics~\cite{raissi2019physicsinformed}, or audio~\cite{gao2021objectfolder}.
Please see \cite{xie2022neuralfields,tewari2021advances} for more details.
In this paper, our focus is on canonicalizing for the 3D pose of neural fields, specifically, neural radiance fields (NeRF)~\cite{mildenhall2020nerf}.

\parahead{Supervised Canonicalization}
Datasets such as ShapeNet~\cite{shapenet2015} and ModelNet40~\cite{wu20153d} have 3D shapes that are manually pre-canonicalized.
This inductive bias aids category-level generalization in problems such as 3D reconstruction~\cite{park2019deepsdf,Occupancy_Networks,tatarchenko2019single,groueix2018papier}.
We can also use these datasets to formulate canonicalization as a supervised learning problem~\cite{wang2019normalized} enabling applications such as 6 degree-of-freedom object pose estimation, multi-view reconstruction~\cite{lei2020pix2surf,sridhar2019multiview}, and reconstruction of articulating~\cite{li2020category,zhang2021strobenet} and non-rigid objects~\cite{zakka2020form2fit}. 
However, our goal is to canonicalize without using manual pose labels.

\parahead{Equivariant Neural Networks}
%
Pose-equivariant networks are equivariant to input pose~\cite{thomas2018tensor,Deng_2021_ICCV,weiler2018learning,weiler20183d} by design.
Some of these methods use Spherical Harmonic functions~\cite{thomas2018tensor,weiler20183d,weiler2018learning,fuchs2020se}, or vector neurons~\cite{Deng_2021_ICCV} to extract equivariant features.
Equivariance is closely related to canonicalization since a canonical pose is also equivariant.
Thus, previous methods have used pose-equivariance for canonicalization~\cite{sajnani2022condor,spezialetti2020learning,sun2020canonical,simeonovdu2021ndf}.
However, these methods have thus far been limited to 3D point clouds, voxels, or meshes.


\parahead{Self-Supervised Canonicalization}
Recent research has shown that self/weak supervision is sufficient for learning pose canonicalization on point clouds~\cite{sajnani2021draco,sajnani2022condor,sun2020canonical, spezialetti2020learning}, 2D key-points~\cite{novotny2019c3dpo}, and images~\cite{Novotny2020C3DM}.
%
None of these previous self-supervised methods can operate directly on neural fields -- to the best of our knowledge, ours is the first.
\section{Background}
\label{sec:background}
We first provide background on the essential components and key ideas behind our method.

\parahead{Neural Radiance Fields (NeRF)}
Given posed RGB images, NeRF and its variants~\cite{mildenhall2020nerf,mildenhall2022nerf} learn a neural network to synthesize novel views of complex scenes.
At inference, for any 3D location and viewing direction, it estimates a direction-independent density value $\sigma$, and a direction-dependent color $c$.
Although designed for novel view synthesis, NeRF implicitly learns the 3D geometry and view-dependent appearance of scenes making it a useful representation of shape and appearance.
While NeRFs excel at fitting a single object instance or scene, it struggles to generalize to object categories partly because objects can be in arbitrary 3D poses -- thus, existing methods use pre-canonicalized datasets~\cite{yu2021pixelnerf}.
In this paper, we provide a way to canonicalize 3D pose without any supervision for objects represented as neural radiance fields.
While our method is aimed at NeRFs, it can be extended to the 3D pose canonicalization of any neural field.

\parahead{Equivariance and Canonicalization}
A function $\Gamma$ over $x$ is said to be \textbf{equivariant} to a group operation $\mathcal{O}$ if its output changes by a fixed
mapping $M$ for any $\mathcal{O}$ operating on the input $x$, \ie,~$\Gamma(\mathcal{O}x) = M(\mathcal{O}) \, \Gamma(x)$.
In this paper, we are interested in equivariance to the $SE(3)$ group denoting 3D position and orientation (pose) -- in particular we focus on the rotation group $SO(3)$ since 3D translation equivariance is readily achieved through mean centering~\cite{novotny2019c3dpo}.
\textbf{Pose canonicalization} is closely related to equivariance -- our goal is to find an \emph{equivariant} canonical 3D rigid transformation that maps the neural field of an object in an arbitrary pose to a canonical pose.
We call the outputs of our method \textbf{canonical fields} -- neural fields that are consistently oriented across a category of shapes.

\begin{figure}[t!]
    \centering
    \includegraphics[width= 0.48\textwidth]{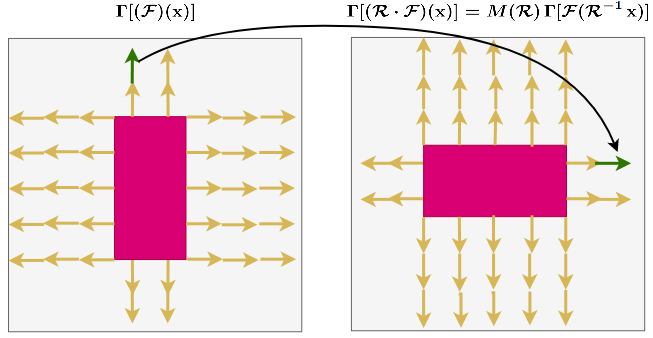}
    \caption{\textbf{Rotation Equivariance in Vector Fields.}
    Rotation equivariance in vector fields $\mathcal{F}$ (here, a simulated 2D magnetic field) require two operations: (1)~update the position of the green arrow (the vector), and (2)~rotating the green arrow at its new position.
    A function $\Gamma$ operating on $\mathcal{F}$ is rotationally equivariant over the field if and only if $\Gamma[(\mathcal{R} \cdot \mathcal{F}) (\mathbf{x})] = M(\mathcal{R}) \, \Gamma[\mathcal{F}(\mathcal{R}^{-1} \, \mathbf{x})]$, where $\mathcal{R}$ is an $SO(2)$ rotation, and $M$ is a rotation-dependent mapping function.
    }
    \label{fig:field_eq}
\end{figure}

\parahead{Rotation Equivariance in Vector Fields}
Unlike 3D point clouds, rotation equivariance in fields is more involved.
Consider a 2D vector field $\mathcal{F}$ defined $\forall \mathbf{x} \in \mathbb{R}^2$ (see \cref{fig:field_eq}).
Rotating this field requires two operations: (1)~updating the positions of vectors in the field to new rotated positions, and (2)~rotating the directions of the vectors.
%
A function $\Gamma$ operating on $\mathcal{F}$ is rotationally equivariant over the field if and only if $\Gamma[(\mathcal{R} \cdot \mathcal{F}) (\mathbf{x})] = M(\mathcal{R}) \, \Gamma[\mathcal{F}(\mathcal{R}^{-1} \, \mathbf{x})]$, where $\mathcal{R}$ is an $SO(2)$ rotation, and $M$ is a rotation-dependent mapping function.
In this paper, we operate on NeRF's density field (a scalar field) but also extract equivariant features that can be vector-valued fields.

\parahead{Tensor Field Networks (TFNs)}
Many methods have been proposed for using equivariant features for rotation canonicalization of 3D data, for instance, spherical CNNs~\cite{cohen2018spherical,spezialetti2020learning}, vector neurons~\cite{Deng_2021_ICCV}, or capsule networks~\cite{zhao20193d,sun2020canonical}.
However, these and other methods~\cite{sajnani2021draco} are limited to 3D point cloud or voxel representations.
In this paper, we extend Tensor Field Networks~\cite{thomas2018tensor,poulenard2021functional} pose canonicalization directly on samples from NeRF's density field.
TFNs operate by computing Type-$\ell$ real-valued spherical harmonic functions that are rotation equivariant: $D^{\ell}(R)Y^{\ell}(X) = Y^{\ell}(Rx)$. 
Where $D^{\ell}: \text{SO}(3) \to \text{SO}(2 \ell + 1)$ is the Wigner matrix of type $\ell$ (please see \cite{thomas2018tensor,poulenard2021functional} for details).  
For a 3D point set $P \in \mathbb{R}^{N \times 3}$, a rotation-equivariant TFN convolution layer (EQConv) at point $p \in P$ is defined as: $\text{EQConv}^{J}(p) = Q^{(n,\ell), J}\left(\sum_{y \in {}^{2R}\mathcal{N}} k^{n}_r(p - y) \otimes s^{\ell}[y]\right)$. 
%
%
Where ${}^{2R}\mathcal{N}$ is a set containing neighbors of point $p$ at twice the resolution and $Q^{(n,\ell), J}(\cdot)$ is the Clebsh Gordon decomposition to combine type-$n$ kernel $k^{n}_r$ and type-$\ell$ equivariant signal $s^{\ell}$.
The $\text{EQConv}(\cdot)$ layer aggregates type-$\ell$ feature $s^{\ell}$ at multiple receptive fields along with learnable weights to perform equivariant convolution.
Our method uses the EQConv layer as the fundamental building block to canonicalize \emph{continuous} fields.

%



\begin{figure*}
    \centering
    \includegraphics[width=0.98\textwidth]{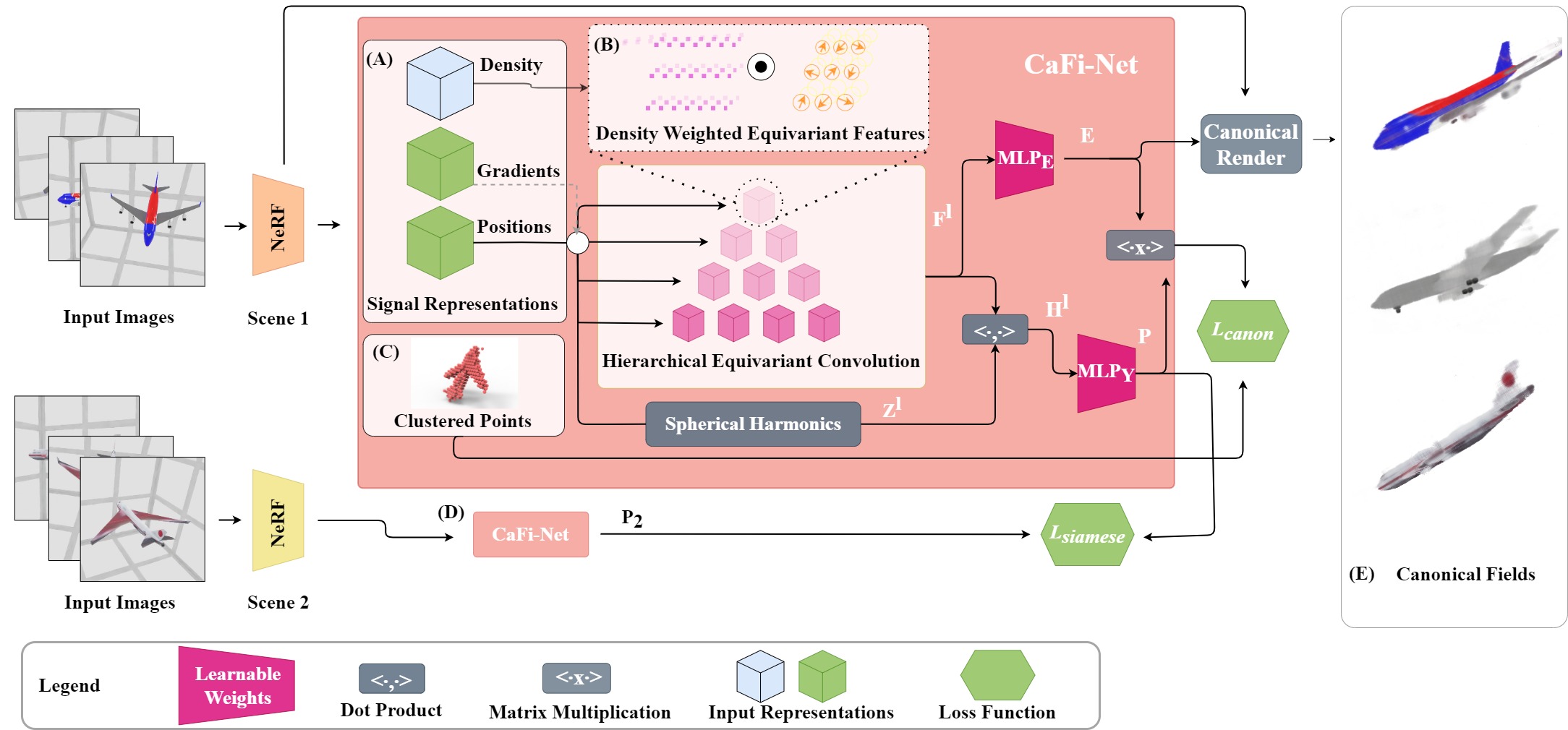}
    \caption{\textbf{\MethodName} samples a density field from NeRF and uses its \textbf{density}, \textbf{position} and \textbf{density gradients} as input signals (A) to canonicalize the field. We predict rotation equivariant features and weigh them by density (B) to guide our learning from the occupied regions of the scene. We then compute an invariant embedding by taking a dot product between equivariant features. This invariant embedding is used to canonicalize the field that enables rendering all the objects in the canonical frame (E). Our method also applies an inter-instance consistency loss (D) that aligns different instances of the same category in the canonical frame. \textbf{We do not assume pre-canonicalized fields, and canonicalize in a self-supervised manner}.}
    \label{fig: Pipeline}
     \vspace{-3.mm}
\end{figure*}
\section{Method}
\label{sec:method}
Our method assumes that we have pre-trained NeRF models of scenes containing single object instances at arbitrary poses from common shape categories.
Our scenes contain primarily the object, but also some background -- perfect foreground-background segmentation is not available.
Our goal is to build a method that estimates a canonicalizing $SE(3)$ transformation that aligns these object NeRFs to a consistent category-level \textbf{canonical field}.

Rather than operate on 3D point clouds, meshes, or voxels, our method (see \cref{fig: Pipeline}) operates directly on samples from the continuous neural radiance field.
During training, we learn \textbf{canonical fields} from a pre-trained NeRF dataset in a fully \textbf{self-supervised} manner.
We use a Siamese network architecture that extracts equivariant field features for canonicalization (see ~\cref{fig: Pipeline}). 
During inference, given a previously unseen NeRF model of a new object instance in an arbitrary pose, our method estimates the transformation that maps it to the canonical field.





\subsection{NeRF Sampling}
For a pre-trained neural radiance field
that maps 3D position $(x,y,z)$ and direction $\theta,\phi$ to color and density $(\textbf{c}, \sigma)$, the input to our method consists of samples on the density field.
We do not use color because it is direction-dependent in NeRF and also varies for different instances.

\parahead{Preprocessing}
During both training and inference, we uniformly sample the pre-trained NeRF density field to find the object center and its bounding box (see supplementary document for details).
We then sample a uniform grid in the density field of the object's bounding box.
We empirically choose uniform grid sampling over fully random sampling since we found no difference.
The uniformly sampled density grid is then normalized to obtain density values $\bm{\sigma}_D$ in the continuous range $[0, 1]$, \ie,~   

\begin{flalign}
 &\bm{\sigma}_D := \{1 - exp(-d \cdot \Gamma^{\sigma}_r(x)) \;|\; x \in \bm{X}\},
\end{flalign}
where $\bm{X}$ is the regularly sampled grid within the object bounding box and $d$ is the depth step size used in the coarse NeRF sampling.
\textbf{Translation} and \textbf{scale} canonicalization is simply achieved by moving the bounding box's center~\cite{novotny2019c3dpo} and scaling it, so our next concern is on rotation canonicalization.

\subsection{\MethodName}
\MethodName (\underline{Ca}nonical \underline{Fi}eld \underline{Net}work) is our method for 3D rotation canonicalization using NeRF's density field (red box in \cref{fig: Pipeline}). We now describe its components.

\parahead{Signal Representation}
At each of the sampled density field locations, we have several choices of signals to use as the input for \MethodName.
For instance, we can (1)~use the NeRF density value directly, (2)~use the $xyz$ location of the sampled field, or (3)~use the gradient of the density field.
Densities alone do not capture the position and directional components of the field, so we use a combination of densities and gradients of densities at query locations of the grid as our input signal for \MethodName. Gradients of density capture the object surface and help in canonicalization (see ablation study in \cref{sec:expt}). 

\parahead{Equivariant Convolution}
To canonicalize for rotation, we leverage equivariance properties of spherical harmonic functions to obtain rotation-equivariant learnable features on the field \cite{thomas2018tensor,poulenard2021functional}.
Type-$\ell$ spherical harmonic functions ($Y^{\ell}$) are degree-$\ell$ polynomials of points on the sphere that are equivariant to $SO(2\ell + 1)$.
The densities sampled from NeRF reconstructions are often noisy and contain outliers.
To promote our method to learn only from occupied regions, we weigh each equivariant feature type by its density.
This weighing has no effect on the equivariance of features (see supplementary document for proof).
We have the option to weigh equivariant features with the local average of the density to smoothen the signal or to weigh directly by the density. We found weighing by density worked better than taking the local average density (see \cref{sec:expt}).
We learn equivariant features of type $\ell$ as:

\resizebox{.95\linewidth}{!}{
  \begin{minipage}{1.1\linewidth}
\begin{flalign}
& {}^{R}s^{\ell} := f_{w}({}^{R}x) \left(W^\ell (\text{EQConv}^\ell(\nabla (\bm{\sigma}_D({}^{R}x)))) + \delta_{0\ell}b\right),\\
&f_{w}({}^{R}x) :=
    \begin{cases}
      \text{mean}_{r} (\bm{\sigma}_D({}^{R}x)), & or\\
      \bm{\sigma}_D({}^{R}x), 
    \end{cases}
\end{flalign}
\end{minipage}
}\\
%
where, $W^{\ell}$ and $b$ are the weights and biases, ${}^{R}x$ is a 3D point queried at resolution $R$ and $\text{mean}_{r} (\bm{\sigma}_D({}^{R}x))$ is the average density at location $^{R}x$ from points sampled at radius $r$.
We hierarchically aggregate features at resolutions $\frac{1}{2}, \frac{1}{4}$, and $\frac{1}{8}$ to obtain global equivariant features. The $\text{EQConv}^{\ell}$ convolution layer is the same as defined in \cref{sec:background}.
We employ non-linearities from \cite{poulenard2021functional} that preserve equivariance and capture better equivariant features. We also compute the max-pool of point-wise features of the last layer to obtain global type-$\ell$ equivariant features $F^{\ell}$.  

\parahead{Canonicalization}
\label{sec:field_canonicalization}
After obtaining the global type-$\ell$ equivariant feature $F^{\ell}$, we compute its dot product with point-wise spherical harmonics $Z^{\ell}$ scaled by their norms $Z^{\ell}(\bm{X}) = \norm{\bm{X}}_2 Y^{\ell}(\bm{X}/\norm{\bm{X}}_2)$ to obtain a pose-invariant embedding $H^{\ell}$ \cite{sajnani2022condor}. As the two vectors $F^{\ell}$ and $Z^{\ell}$ are both equivariant to input rotation, their dot product is invariant. We use a linear layer on top of the pose-invariant embedding to estimate grid coordinates in the canonical frame ($P \in \mathbb{R}^{H \times W \times D \times 3}$) for each point in $\bm{X}$ and $M$ equivariant transformations $E \in \mathbb{R}^{M \times 3 \times 3}$ \cite{sajnani2022condor} that orient the canonical coordinates to the input coordinates. We then choose the best canonicalizing transform $E_{b}$ and penalize it for canonicalization.
The invariant embedding $H$, grid coordinates $P$ and equivariant rotation $E$ are given as,
%
\begin{flalign}
    & H^{\ell}(\bm{\sigma}_D, \bm{X}) := \langle F^{\ell}(\bm{\sigma}_D), Z^{\ell}(\bm{X}) \rangle, \\
    & P := \text{MLP}_{Y}(H(\bm{\sigma}_D,\bm{X})), \\
    & E := \text{MLP}_{E}(F(\bm{\sigma}_D,\bm{X})).
\end{flalign}
Here, we have dropped the type-$\ell$ notation for $H$ and $F$ as we concatenate all equivariant feature types. 

\parahead{Siamese Network Architecture}
Experimentally, canonicalizing noisy density fields is difficult without guidance on shape similarity within a category.
We therefore use a Siamese training strategy~\cite{novotny2019c3dpo} where two different object instances are forced to be consistently canonicalized (see the lower branch in \cref{fig: Pipeline}).

\parahead{Clustering}
\MethodName predicts a canonical coordinate for every grid point, but unlike point clouds, fields have many unoccupied regions that do not provide any information.
We penalize the network on foreground regions only by performing K-means clustering on the densities within the object bounding box with $K = 2$ and choose the cluster $\bm{C}_f$ with a higher mean.
Note that we still operate on continuous fields, but only cluster samples to guide our training. 


\subsection{Training}
\MethodName is trained on a large dataset of pre-trained NeRF models over 13 object categories (see \cref{sec:expt}).
We use the following loss functions to train our model.


\parahead{Canonicalization Loss}
To self-supervise our learning, we transform the predicted canonical grid coordinates to the input grid using the canonical rotation $E_b$ and compute a point-wise L2 distance between them.
This loss forces the predicted invariant grid coordinates $P$ to reconstruct the shape and forces the predicted canonicalizing transform to be equivariant to the input transformation. The final loss is computed over the canonicalizing transform that minimizes this loss:
\begin{flalign}
    & E_b := \min_{j} \left(\text{mean}_{i \in \bm{C}_f} \norm{\bm{X}_i - E_j P_i}_2^{2}\right),\\
    & \mathcal{L}_{canon} := \text{mean}_{i \in \bm{C}_f} \norm{\bm{X}_i - E_b P_i}_2^2,
\end{flalign}
where $C_f$ is a set of points belonging to the foreground.

\parahead{Orthonormality Loss}
The predicted equivariant transformations $E$ should have orthonormal vectors and we use the following to force orthonormality using:
\begin{flalign}
     & U_j, D_j, V_j :=  \text{SVD}(E_j),\\&
     \mathcal{L}_{ortho} := \frac{1}{M} \sum_j \norm{E_j - U_j V^{T}_j}_2,
\end{flalign}
where $\text{SVD}(E_j)$ computes the singular value decomposition of the $j^{th}$ equivariant transformation.

\parahead{Siamese Shape Loss}
We use a Siamese loss that penalizes for consistency between different shapes belonging to the same category.  Given two un-canonicalized fields in different frames of reference $\bm{\sigma}^{1}_D$ and $\bm{\sigma}^{2}_D$, we canonicalize both the fields to the canonical frame and compute chamfer distance between their predicted canonical coordinates for points in the foreground cluster $C_f$, \ie,
\begin{eqnarray}
    \mathcal{L}_{siamese} := \text{CD}(P(\bm{\sigma}^{1}_D), P(\bm{\sigma}^{2}_D)).
\end{eqnarray}
This loss regularizes the training by ensuring that instances of the same category should be aligned in the canonical frame.



\parahead{Architecture and Hyper-parameters}
%
\MethodName predicts a $128$ dimensional invariant embedding for each point in space and performs convolution at resolutions $1/2$, $1/4$, and $1/8$. Each equivariant convolution layer aggregates features from ~$512$ neighboring points at twice the resolution.  We then use equivariant non-linearities introduced in \cite{poulenard2021functional} by applying Batch Normalization \cite{batchnorm} and ReLU activation \cite{relu} in the inverse spherical harmonic transform domain. We use a three-layer MLP with intermediate Batch Normalization and ReLU layers followed by a final linear layer to predict canonical coordinates. To train our network, we weigh the canonicalization loss ($\mathcal{L}_{canon}$) the highest with weight $2.0$ and $1.0$ for all the other loss functions.
All our models are written in PyTorch~\cite{PyTorch} and are trained for 300 epochs with a batch size of $2$ on an Nvidia 1080-Ti GPU. We use the Adam Optimizer~\cite{Adam} with an initial learning rate of $6e-4$ and L2 weight regularization of $1e-5$ for all our experiments.
Please see the supplementary document for more details.

    \section{Experiments}
\label{sec:expt}

\begin{table*}[!ht]
\centering
    \caption{This table compares the canonicalization performance of our method (F - operating on fields) with other 3D point cloud-based methods (P) on three standard metrics (IC, CC and GEC) on our dataset of 13 categories.
    We compare with PCA, Canonical Capsules (CaCa)~\cite{sun2020canonical} and ConDor~\cite{sajnani2022condor}.
    All metrics are multiplied by 100 for ease of reading. The top two performing methods are highlighted in \textbf{{\color{magenta}{magenta}}} (best) and \textbf{{\color{blue}{\textbf{blue}}}} (second best).
    We are better than SOTA~\cite{sajnani2022condor} on the Ground Truth Equivariance Consistency (GEC) and Category-Level Consistency (CC) with lower mean and median canonicalization error. However, we perform on par with ConDor~\cite{sajnani2022condor} on the Instance-Level Consistency.
    \vspace{-0.1in}
    \label{table:canonicalization_metrics_full}
    }
\scalebox{0.75}{
\begin{tabular}{r|rrrrrrrrrrrrr|rr}
    \toprule
     & \textbf{bench} & \textbf{cabinet} & \textbf{car} & \textbf{cellph.} & \textbf{chair} & \textbf{couch} & \textbf{firearm} & \textbf{lamp} & \textbf{monitor} & \textbf{plane} & \textbf{speaker} & \textbf{table} & \textbf{water.} & \textbf{avg.} & \textbf{med.}  \\
     
     \midrule
    \multicolumn{16}{l}{\textbf{Instance-Level Consistency (IC)} {\color{blue} $\downarrow$}} \\
    
    \midrule
    \multicolumn{1}{l|}{PCA (P)} & {14.65} & {5.94} & {6.13} & \textbf{\color{blue} 0.67} & 6.13 & 7.63 & 15.07 & 12.84 & {6.78} & 7.90 & {5.40} & {10.31} & {10.72} & 8.47 & 7.63\\
    \multicolumn{1}{l|}{CaCa (P)~\cite{sun2020canonical}} & 2.89 & 2.09 & 2.01 & 2.81 & 1.08 & 1.90 & \textbf{\color{magenta} 0.23} & \textbf{\color{magenta} 2.97} & 2.35 & 1.51 & 2.21 & 2.42 & \textbf{\color{blue} 2.60} & 2.08 & 2.21 \\
    \multicolumn{1}{l|}{ConDor (P)~\cite{sajnani2022condor}} & \textbf{\color{magenta} 0.56} & \textbf{\color{blue} 1.27} & \textbf{\color{magenta} 0.36} & \textbf{\color{magenta} 0.53} & \textbf{\color{blue} 0.96} & \textbf{\color{magenta} 0.32} & \textbf{\color{blue} 0.60} & \textbf{\color{blue} 3.78} & \textbf{\color{blue} 0.44} & \textbf{\color{blue} 0.64} & \textbf{\color{blue} 1.97} & \textbf{\color{blue} 1.32} & \textbf{\color{magenta} 2.15} & \textbf{\color{magenta} 1.15} & \textbf{\color{magenta} 0.64} \\ 
    \midrule
    
    \multicolumn{1}{l|}{Ours (F)} & \textbf{\color{blue} 2.62} & \textbf{\color{magenta} 1.15} & \textbf{\color{blue} 0.39} & 1.28 & \textbf{\color{magenta} 0.81} & \textbf{\color{blue} 1.43} & 0.74 & 3.87 & \textbf{\color{magenta} 0.41} & \textbf{\color{magenta} 0.36} & \textbf{\color{magenta} 0.60} & \textbf{\color{magenta} 1.09} & 2.74 & \textbf{\color{blue} 1.34} & \textbf{\color{blue} 1.09} \\ 
    \midrule
    
    \multicolumn{16}{l}{\textbf{Category-Level Consistency (CC)} {\color{blue} $\downarrow$}} \\
    
    \midrule
    \multicolumn{1}{l|}{PCA (P)} & {14.55} & 9.68 & {7.48} & 4.26 & 9.90 & 13.45 & 15.13 & {11.87} & 13.09 & {8.15} & {5.51} & {10.43} & {11.07} & 10.36 & 10.43 \\
    \multicolumn{1}{l|}{CaCa (P)~\cite{sun2020canonical}} & 2.79 & \textbf{\color{magenta} 0.38} & 1.05 & \textbf{\color{magenta} 0.83} & 1.82 & \textbf{\color{blue} 1.59} & 1.31 & \textbf{\color{blue} 4.34} & \textbf{\color{magenta} 0.39} & 1.83 & \textbf{\color{magenta} 0.65} & \textbf{\color{blue} 1.91} & \textbf{\color{magenta} 2.06} & \textbf{\color{blue} 1.61} & 1.59 \\
    \multicolumn{1}{l|}{ConDor (P)~\cite{sajnani2022condor}} & \textbf{\color{magenta} 1.98} & 1.68 & \textbf{\color{magenta} 0.47} & \textbf{\color{blue} 1.04} & \textbf{\color{blue} 1.28} & \textbf{\color{magenta} 0.88} & \textbf{\color{blue} 0.96} & 4.44 & 1.58 & \textbf{\color{blue} 1.26} & 2.10 & 2.10 & 2.56 & 1.72 & \textbf{\color{blue} 1.58} \\
    \midrule
    \multicolumn{1}{l|}{Ours (F)} & \textbf{\color{blue} 2.77} & \textbf{\color{blue} 1.36} & \textbf{\color{blue} 0.48} & 1.26 & \textbf{\color{magenta} 0.75} & \textbf{\color{blue} 1.59} & \textbf{\color{magenta} 0.92} & \textbf{\color{magenta} 3.92} & \textbf{\color{blue} 0.58} & \textbf{\color{magenta} 0.63} & \textbf{\color{blue} 0.77} & \textbf{\color{magenta} 1.48} & \textbf{\color{blue} 2.35} & \textbf{\color{magenta} 1.45} & \textbf{\color{magenta} 1.26} \\
    \midrule
    
    \multicolumn{16}{l}{\textbf{Ground Truth Equivariance Consistency (GEC)}{\color{blue} $\downarrow$}} \\ 
    
    \midrule
    \multicolumn{1}{l|}{PCA (P)} & 14.40 & {5.86} & 6.32 & 1.64 & 6.32 & 8.12 & 15.01 & 12.49 & 6.82 & 8.16 & {5.32} & 10.17 & {11.27} & 8.61 & 8.12\\
    \multicolumn{1}{l|}{CaCa (P)~\cite{sun2020canonical}} & 3.65 & 2.28 & 2.40 & 2.74 & 1.90 & 2.22 & 1.28 & 4.69 & 2.48 & 2.09 & \textbf{\color{blue} 2.22} & 3.01 & \textbf{\color{blue} 2.79} & 2.59 & 2.40 \\ 
    \multicolumn{1}{l|}{ConDor (P)~\cite{sajnani2022condor}} & \textbf{\color{magenta} 2.12} & \textbf{\color{blue} 1.79} & \textbf{\color{magenta} 0.50} & \textbf{\color{magenta} 1.17} & \textbf{\color{blue} 1.34} & \textbf{\color{magenta} 0.93} & \textbf{\color{blue} 1.04} & \textbf{\color{blue} 4.58} & \textbf{\color{blue} 1.59} & \textbf{\color{blue} 1.31} & \textbf{\color{blue} 2.22} & \textbf{\color{blue} 2.24} & \textbf{\color{magenta} 2.67} & \textbf{\color{blue} 1.81} & \textbf{\color{blue} 1.59} \\
    \midrule
    \multicolumn{1}{l|}{Ours (F)} & \textbf{\color{blue} 3.26} & \textbf{\color{magenta} 1.51} & \textbf{\color{blue} 0.54} & \textbf{\color{blue} 1.43} & \textbf{\color{magenta} 0.94} & \textbf{\color{blue} 1.90} & \textbf{\color{magenta} 1.01} & \textbf{\color{magenta} 4.37} & \textbf{\color{magenta} 0.63} & \textbf{\color{magenta} 0.66} & \textbf{\color{magenta} 0.81} & \textbf{\color{magenta} 1.63} & 3.02 & \textbf{\color{magenta} 1.67} & \textbf{\color{magenta} 1.43} \\
    \bottomrule
\end{tabular}}
\vspace{-0.3cm}
\end{table*}

We now provide details on our extensive experiments to evaluate the quality of our pose canonicalization.
To our knowledge, we are the first method for pose canonicalization of neural radiance fields which makes direct comparisons with other work challenging.
Nonetheless, we have designed experiments to compare our method with three 3D point cloud-based canonicalization methods.
Overall, we present three classes of experiments: (1)~evaluation of the performance of our method on 13 different categories, (2)~comparison with other work, and (3)~ablations to justify method design choices.



\parahead{Dataset}
All our experiments use a new large synthetic NeRF dataset of shapes from 13 categories (that overlap with \cite{groueix2018papier,sun2020canonical,sajnani2022condor}) from the ShapeNet~\cite{shapenet2015} dataset.
For each category, we pick 100 instances from ShapeNet, rotate the shape randomly, and render 54 omnidirectional views sampled in a cube around the shape using Blender.
Different from other datasets, we also simulate cluttered backgrounds.
These posed views are used to train 100 NeRF models per category using the public PyTorch implementation of NeRF~\cite{lin2020nerfpytorch} for a total of \textbf{1300 NeRF models}.
To our knowledge, this is one of the largest (compared to \cite{gao2021objectfolder,vora2021nesf}) 360$^\circ$ synthetic NeRF datasets -- we will publicly release the simulator, raw posed images, and trained NeRF weights.
Each NeRF model is trained for 400 epochs with 1024 randomly selected rays per iteration, a coarse sampling resolution of 64 points, and a fine sampling resolution of 128 points along each ray.
Rather than train NeRF to maximize PSNR on a novel view, we fix the number of epochs for all models to enable a fair comparison between instances.
As a consequence, some NeRF models may have a lower PSNR, but \textbf{our goal is not to increase NeRF quality but rather improve canonicalization quality} -- hence we fix the number of epochs.
For each category, we set aside 20\% of the models for testing.
\parahead{Metrics}
Since there are no known metrics for evaluating canonical fields, we resort to 3 metrics used for 3D point cloud canonicalization~\cite{sajnani2022condor}:
(1)~\textbf{Instance-Level Consistency (IC)}, a metric that measures how consistently we can canonicalize the same instance in different poses,
(2)~\textbf{Category-Level Consistency (CC)} to measure how consistently we canonicalize across different instances in the same category, and
(3)~\textbf{Ground Truth Equivariance Consistency (GEC)}, a variant of the Ground Truth Consistency (GC) metric proposed in \cite{sajnani2022condor} to measure canonicalization performance compared to manual labels.
We use the GEC because we observed that the original GC metric is prone to degeneracy when the canonicalizing transform is identity.
We fix this issue by taking three point clouds $P_i, P_j$, and $P_k$ from a canonicalized dataset $\mathcal{P}$ and rotating $P_i$ by $R_1$, $P_j$ by $R_2$ where both $R_1$ and $R_2$ are random rotations.
Let $\mathcal{C}(P)$ predict a canonicalizing rotation for a point cloud $P$, we then compute the Ground Truth Equivariance consistency as:
\resizebox{.85\linewidth}{!}{
  \begin{minipage}{\linewidth}
\begin{eqnarray}
\text{GEC} := \frac{1}{|\mathcal{P}|^3} \sum_{P_i, P_j, P_k \in \mathcal{P}}\text{CD}(\mathcal{C}(R_1P_i)R_1 P_k, \mathcal{C}(R_2P_j)R_2 P_k),
\end{eqnarray}
\end{minipage}
}\\
where CD refers to the Chamfer distance.
Here, we apply canonicalizing transforms of shape $P_i$ and $P_j$ to the shape $P_k$.
This modified metric will not be degenerate for identity canonicalizing transforms and is more rigorous in evaluating ground truth consistency.
To compute our metrics, we used point clouds sampled from the Shapenet meshes~\cite{shapenet2015} using poisson disk sampling.

\begin{table*}
\subcaptionbox{\textbf{Choice of Signal Representation} - Canonicalization metrics for using Gradients vs. $xyz$ locations as input signal. Gradients capture the object surface that help in canonicalization.}{
\scalebox{0.7}{
\begin{tabular}{c|cc}
\toprule
\multicolumn{1}{c|}{{\textbf{Category}}}                          & \multicolumn{2}{c}{\textbf{$xyz$ vs. Gradient Signals}}                                  \\ \cmidrule{1-3} 
\multicolumn{1}{c|}{}                                                          & \multicolumn{1}{c}{\textbf{$xyz$}} & \multicolumn{1}{c}{\textbf{Gradient signal}}\\ \midrule
\multicolumn{3}{l}{\textbf{Ground Truth Equivariance Consistency (GEC)}{\color{blue} $\downarrow$}} \\
\midrule
\multicolumn{1}{l|}{\textbf{bench}}            & \textbf{2.17}                   & 3.26                              \\ 
\multicolumn{1}{l|}{\textbf{cellphone}}            & 1.84                 & \textbf{1.43}                    \\ 
\multicolumn{1}{l|}{\textbf{chair}}            & 1.72                  & \textbf{0.94}                   \\ 
\multicolumn{1}{l|}{\textbf{plane}}            &2.07               &  \textbf{0.66}                  \\ \midrule 
\multicolumn{1}{l|}{Average} & 1.95          & \textbf{1.57}                     \\ \midrule
\multicolumn{3}{l}{\textbf{Instance-Level Consistency (IC)}{\color{blue} $\downarrow$}} \\
\midrule
\multicolumn{1}{l|}{\textbf{bench}}            & \textbf{2.04}                   & 2.62                   \\ 
\multicolumn{1}{l|}{\textbf{cellphone}}            & 1.6                   & \textbf{1.28}       \\ 
\multicolumn{1}{l|}{\textbf{chair}}            &1.4                   & \textbf{0.81}                    \\ 
\multicolumn{1}{l|}{\textbf{plane}} & 1.88         & \textbf{0.36}                     \\ \midrule
\multicolumn{1}{l|}{Average} & 1.73          & \textbf{1.26}                      \\ \midrule

\multicolumn{3}{l}{\textbf{Category-Level Consistency (CC)}{\color{blue} $\downarrow$}} \\
\midrule
\multicolumn{1}{l|}{\textbf{bench}}            & \textbf{1.99}                  & 2.77            \\ 
\multicolumn{1}{l|}{\textbf{cellphone}}            & 1.49                  & \textbf{1.26}                 \\ 
\multicolumn{1}{l|}{\textbf{chair}}            & 1.52                  & \textbf{0.75}                   \\ 
\multicolumn{1}{l|}{\textbf{plane}} & 1.9          & \textbf{0.63}                   \\ \midrule
\multicolumn{1}{l|}{\textbf{Average}}   & 1.72                  & \textbf{1.35}          \\ \midrule
\end{tabular}

}}
\hfill
\subcaptionbox{\textbf{Weighing Equivariant Signals by Local Average Density} deteriorates the performance by smoothing out important details of the shape. We show canonicalization with (\textbf{w}) and without (\textbf{w/o}) weighing by the local averaged density.}{
\scalebox{0.7}{
\begin{tabular}{c|cc}
\toprule
\multicolumn{1}{c|}{{\textbf{Category}}}                          & \multicolumn{2}{c}{\textbf{Local Average Density}}                                  \\ \cmidrule{1-3} 
\multicolumn{1}{c|}{}                                                          & \multicolumn{1}{c}{\textbf{w/o}} & \multicolumn{1}{c}{\textbf{w}}\\ \midrule
\multicolumn{3}{l}{\textbf{Ground Truth Equivariance Consistency (GEC)}{\color{blue} $\downarrow$}} \\
\midrule
\multicolumn{1}{l|}{\textbf{bench}}            & \textbf{3.26}                  & 3.38                                \\ 
\multicolumn{1}{l|}{\textbf{cellphone}}            & \textbf{1.43}                  & 1.63                  \\ 
\multicolumn{1}{l|}{\textbf{chair}}            & \textbf{0.94}                  & 1.3                  \\ 
\multicolumn{1}{l|}{\textbf{plane}}            & 0.66                  & \textbf{0.64}                 \\ \midrule 
\multicolumn{1}{l|}{\textbf{Average}} & \textbf{1.57}          & 1.73                      \\ \midrule
\multicolumn{3}{l}{\textbf{Instance-Level Consistency (IC)}{\color{blue} $\downarrow$}} \\
\midrule
\multicolumn{1}{l|}{\textbf{bench}}            & \textbf{2.62}                   & 2.72                 \\ 
\multicolumn{1}{l|}{\textbf{cellphone}}            & \textbf{1.28}                   & 1.63       \\ 
\multicolumn{1}{l|}{\textbf{chair}}            &\textbf{0.81}                   & 2.64                   \\ 
\multicolumn{1}{l|}{\textbf{plane}} & \textbf{0.36}          & 0.39                    \\ \midrule
\multicolumn{1}{l|}{\textbf{Average}} & \textbf{1.26}          & 1.85                      \\ \midrule

\multicolumn{3}{l}{\textbf{Category-Level Consistency (CC)}{\color{blue} $\downarrow$}} \\
\midrule
\multicolumn{1}{l|}{\textbf{bench}}            & \textbf{2.77}                   & 2.82            \\ 
\multicolumn{1}{l|}{\textbf{cellphone}}            & \textbf{1.26}                  & 1.54                  \\ 
\multicolumn{1}{l|}{\textbf{chair}}            & \textbf{0.75}                  & 1.07                   \\ 
\multicolumn{1}{l|}{\textbf{plane}} & 0.63          & \textbf{0.51}                  \\ \midrule
\multicolumn{1}{l|}{\textbf{Average}}   & \textbf{1.35}                  & 1.49         \\ \midrule
\end{tabular}
}}
\hfill
\subcaptionbox{\textbf{Siamese Training} improves performance on all canonicalization metrics on average. We show canonicalization performance \textit{with siamese} and without \textbf{(w/o) siamese} training.
The average of Ground Truth Equivariance Consistency $GEC$ metric reduces to $1.57$ from $1.86$}{
\scalebox{0.7}{
\begin{tabular}{c|cc}
\toprule

\multicolumn{1}{c|}{{\textbf{Category}}}                          & \multicolumn{2}{c}{\textbf{Siamese Training}}                                  \\ \cmidrule{1-3} 
\multicolumn{1}{c|}{}                                                          & \multicolumn{1}{c}{\textbf{w/o siamese}} & \multicolumn{1}{c}{\textbf{with siamese}}\\ \midrule
\multicolumn{3}{l}{\textbf{Ground Truth Equivariance Consistency (GEC)}{\color{blue} $\downarrow$}} \\
\midrule
\multicolumn{1}{l|}{\textbf{bench}}            & 3.52                  & \textbf{3.26}                                 \\ 
\multicolumn{1}{l|}{\textbf{cellphone}}            & 1.61                   & \textbf{1.43}                   \\ 
\multicolumn{1}{l|}{\textbf{chair}}            & 1.02                  & \textbf{0.94}                  \\ 
\multicolumn{1}{l|}{\textbf{plane}}            & 1.31                  & \textbf{0.66}                   \\ \midrule
\multicolumn{1}{l|}{\textbf{Average}} & 1.86          & \textbf{1.57}                     \\ \midrule
\multicolumn{3}{l}{\textbf{Instance-Level Consistency (IC)}{\color{blue} $\downarrow$}} \\
\midrule
\multicolumn{1}{l|}{\textbf{bench}}         & 2.73                   & \textbf{2.62}                   \\ 
\multicolumn{1}{l|}{\textbf{cellphone}}            & 1.3                 & \textbf{1.28}           \\ 
\multicolumn{1}{l|}{\textbf{chair}}      & 0.82                  & \textbf{0.81}                    \\ 
\multicolumn{1}{l|}{\textbf{plane}}           & 1.1                  & \textbf{0.36}                   \\ \midrule 
\multicolumn{1}{l|}{\textbf{Average}} & 1.48          & \textbf{1.26}                  \\ \midrule
\multicolumn{3}{l}{\textbf{Category-Level Consistency (CC)}{\color{blue} $\downarrow$}} \\
\midrule
\multicolumn{1}{l|}{\textbf{bench}}            & 2.95                   & \textbf{2.77}            \\ 
\multicolumn{1}{l|}{\textbf{cellphone}}            & 1.44                   & \textbf{1.26}                 \\ 
\multicolumn{1}{l|}{\textbf{chair}}            & 0.83                  &\textbf{0.75}                 \\ 
\multicolumn{1}{l|}{\textbf{plane}}            & 1.26                  & \textbf{0.63}                           \\ \midrule
\multicolumn{1}{l|}{\textbf{Average}}   & 1.62                  & \textbf{1.35}          \\ \midrule
\end{tabular}
}}
\vspace{-0.2cm}
\caption{Ablation studies to justify three key design choices: (a)~signal representation, (b)~density weighting, and (c)~Siamese architecture.}
\label{tab:ablation}
\end{table*}
\vspace{-0.2cm}


%
\subsection{Evaluation \& Comparisons}
We first analyze the performance of our method on the three metrics across 13 different categories.
In \cref{table:canonicalization_metrics_full}, we highlight our method that operates on a field as \textbf{Ours (F)}.
Our method performs well across all categories with low variance between different categories.
We observe similar trends across metrics for a set of categories (\eg,~firearms,lamps) indicating that category shape plays a role in canonicalization performance. 

Next, we compare our method against three benchmarks: Principal Component Analysis (PCA)~\cite{pearson1901liii}, Canonical Capsules (CaCa)~\cite{sun2020canonical}, and ConDor~\cite{sajnani2022condor}.
All prior methods canonicalize clean point clouds with all points on the surface.
To have a fair comparison, we train the above methods on point clouds (P) sampled from NeRF scenes for points on and within the object. Most scenes have points in the range of 120--1200 when sampled uniformly on the $32^3$ grid.
We resample the scene when points are less than 1024 and train both Canonical Capsules and ConDor on the resulting point clouds. 
\MethodName estimates a canonicalizing transform directly from the density field that we use to transform its corresponding ShapeNet point cloud for comparison against point canonicalizers. We follow the same protocol as \cite{sajnani2022condor} to measure and benchmark canonicalization performance for all the methods. 

\begin{figure*}[t!]
\centering
  \includegraphics[width=\textwidth]{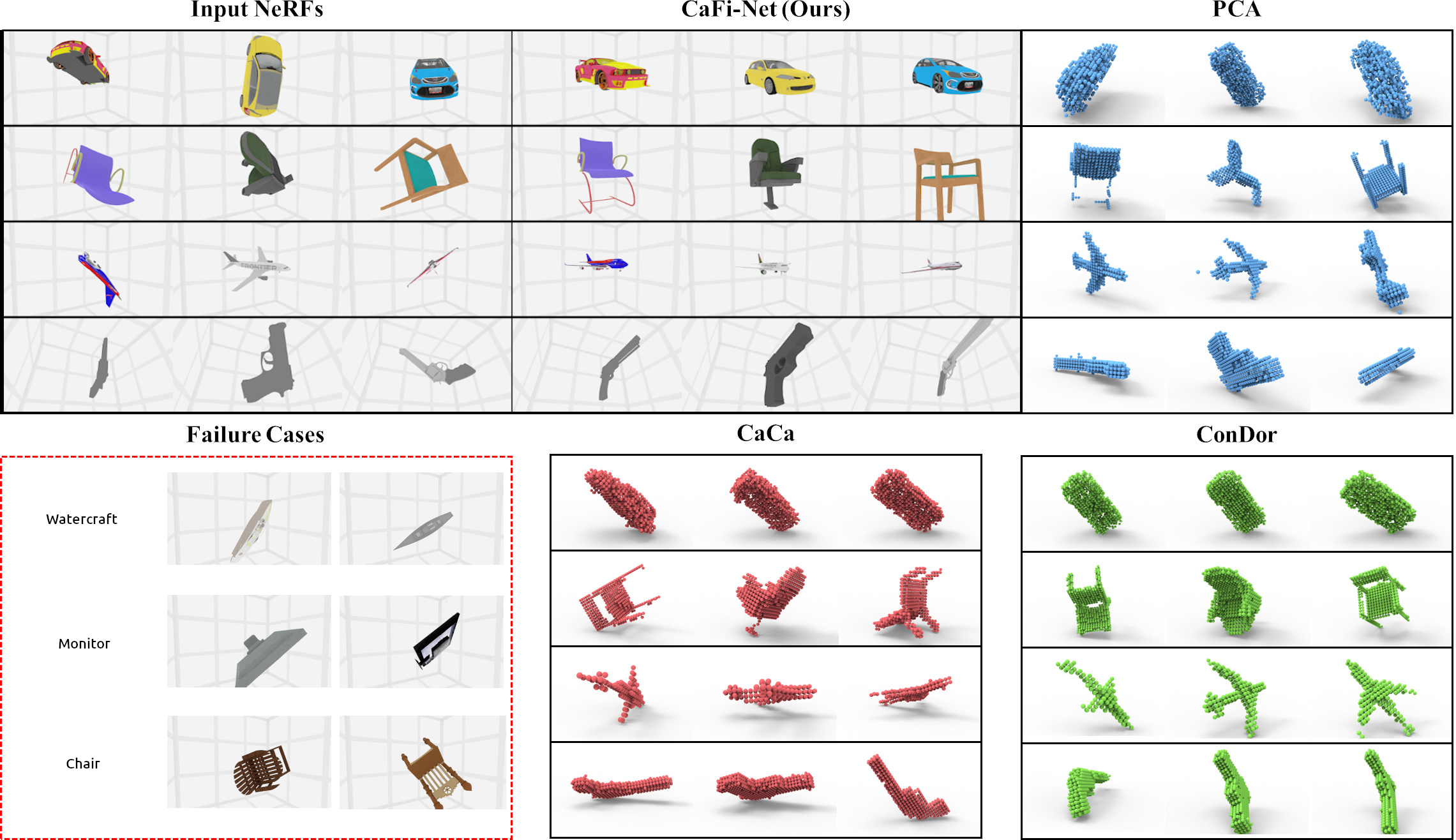}
  \vspace{-0.3in}
  \caption{
    This figure visually compares the canonicalization performance of our method with 3D point cloud-based methods.
    \textbf{Input NeRFs} shows one of the RGB views used to learn our NeRF input (notice how orientations of instances are random).
    \textbf{CaFi-Net (Ours)} shows the results of our canonical field rendered from a novel view unseen during NeRF training.
    \textbf{PCA} (blue), \textbf{CaCa}~\cite{sun2020canonical} (red), and \textbf{ConDor}~\cite{sajnani2022condor}'s (green) results are also presented for the same instances.
    Our method matches or exceeds ConDor.
    We also show some failure cases for several instances (red box) which occur either due to thin structures or symmetry.
  }
  \label{fig:money_shot}
  \vspace{-0.2in}
\end{figure*}

\parahead{Analysis}
\cref{table:canonicalization_metrics_full} compares the metrics for PCA (P), CaCa (P), ConDor (P), and \MethodName [Ours (F)].
We list point cloud canonicalizers with (P) and field canonicalizer as (F) in the table.
On average, our method performs better than point cloud-based methods in the CC and GEC metrics and is on par with the state-of-the-art method (ConDor) in the IC metric.
This is despite the fact that our method directly operates on the field. 
%
We observe that PCA underperforms in all categories suggesting that the noisy point clouds make it unreliable.
Similarly we observe that CaCa fails to canonicalize in categories where high noise is observed.
Given that ConDor also uses TFNs in a different architecture, we conclude that the choice of TFNs makes canonicalization more robust to noise.
Our method has the additional advantage of operating directly on the field.
Please see \cref{fig:money_shot} for a qualitative comparison of the different methods.
%



\subsection{Ablations}

We conduct additional experiments to justify key design choices in our method.
For all the ablation studies we use a smaller subset of our data consisting of 4 categories: bench, cellphone, chair, and airplane.

\parahead{Choice of Signal Representation}
To select the appropriate signal representation to use (see \cref{sec:method}), we conducted an ablation study using just the density field $xyz$ locations or its gradients. 
As seen in \cref{tab:ablation}~(a), the average canonicalization error over 4 categories is lower when using gradient as compared to using density $xyz$ locations on all three metrics. We hypothesize that gradient of density weighs equivariant features closer to the surface of the object higher. This improves the canonicalization performance. 

\parahead{Weighing Features by Local Average Density}
Next, we justify why weighting the equivariant features using the density is helpful.
\cref{tab:ablation} (b) shows results for weighting features by local average density compared to weighting features by density.
Weighing features by local average smoothens out details that are necessary to canonicalize the volume. Our average ground truth canonicalization error increases from $1.57$ to $1.73$ if we weigh features by local averaged densities. In future works, we will delve into better methods for reducing input signal noise that does not smoothen out the geometry details of objects to further improve our canonicalization.

\parahead{Siamese Architecture}
Finally, we justify the need for a two-branch Siamese network architecture.
Although our method can canonicalize even without this architecture, it makes it easier to learn over instances in a category as observed in \cite{novotny2019c3dpo}.
\cref{tab:ablation} (c) compares our method on using a single branch architecture and a Siamese architecture.
Clearly, the Siamese architecture helps improve results confirming our hypothesis. Enforcing different instances of the same category to align with each other reduces the Ground Truth Canonicalization error from $1.86$ to $1.57$. We also see a decrease in all the other metrics including Instance-Level Consistency. 
\section{Conclusion}
\label{sec:end}
We presented Canonical Field Network (\MethodName), a method for self-supervised category-level canonicalization of the 3D pose of objects represented as neural radiance fields.
\MethodName consists of a Siamese network architecture with rotation-equivariant convolution layers to extract features for pose canonicalization.
Our approach operates directly on NeRF's noisy but continuous density field.
We train our method on a large dataset of 1300 NeRF models obtained for 13 common ShapeNet categories.
During inference, our method is able to canonicalize arbitrarily oriented NeRFs.
Experiments show that our method matches or outperforms 3D point cloud-based methods.

\parahead{Limitations \& Future Work}
Our approach has several limitations that future work should investigate.
First, we require a 360$^\circ$ NeRF model of objects and cannot handle partial views from front-facing NeRFs. Second, canonicalizing densities with uncertainty is more difficult as the uncertainty increases for very thin/small structures that can be missed (see chair in \cref{fig:money_shot}). Furthermore, we also inherit the issues that TFNs have with symmetry (see monitor case in ~\cref{fig:money_shot}).
In future work, we plan to extend our approach to create a large real dataset of common object categories without requiring manual pose canonicalization and combat the symmetry issue in \MethodName as in \cite{seo2022equisym}\\ 
\parahead{Acknowledgments}
This work was supported by AFOSR grant FA9550-21-1-0214, NSF grant CNS-2038897, an AWS Cloud Credits award, NSF CloudBank, and a gift from Meta Reality Labs. We thank Chandradeep Pokhariya and Ishaan Shah.

{\small
\bibliographystyle{ieee_fullname}
\bibliography{references}
}


\appendix
\newtheorem{theorem}{Theorem}
\newtheorem{lemma}[theorem]{Lemma}

\begin{figure*}[ht]
    \centering
    \includegraphics[trim={0cm 0cm 0cm 0},clip,width=1.0\textwidth]{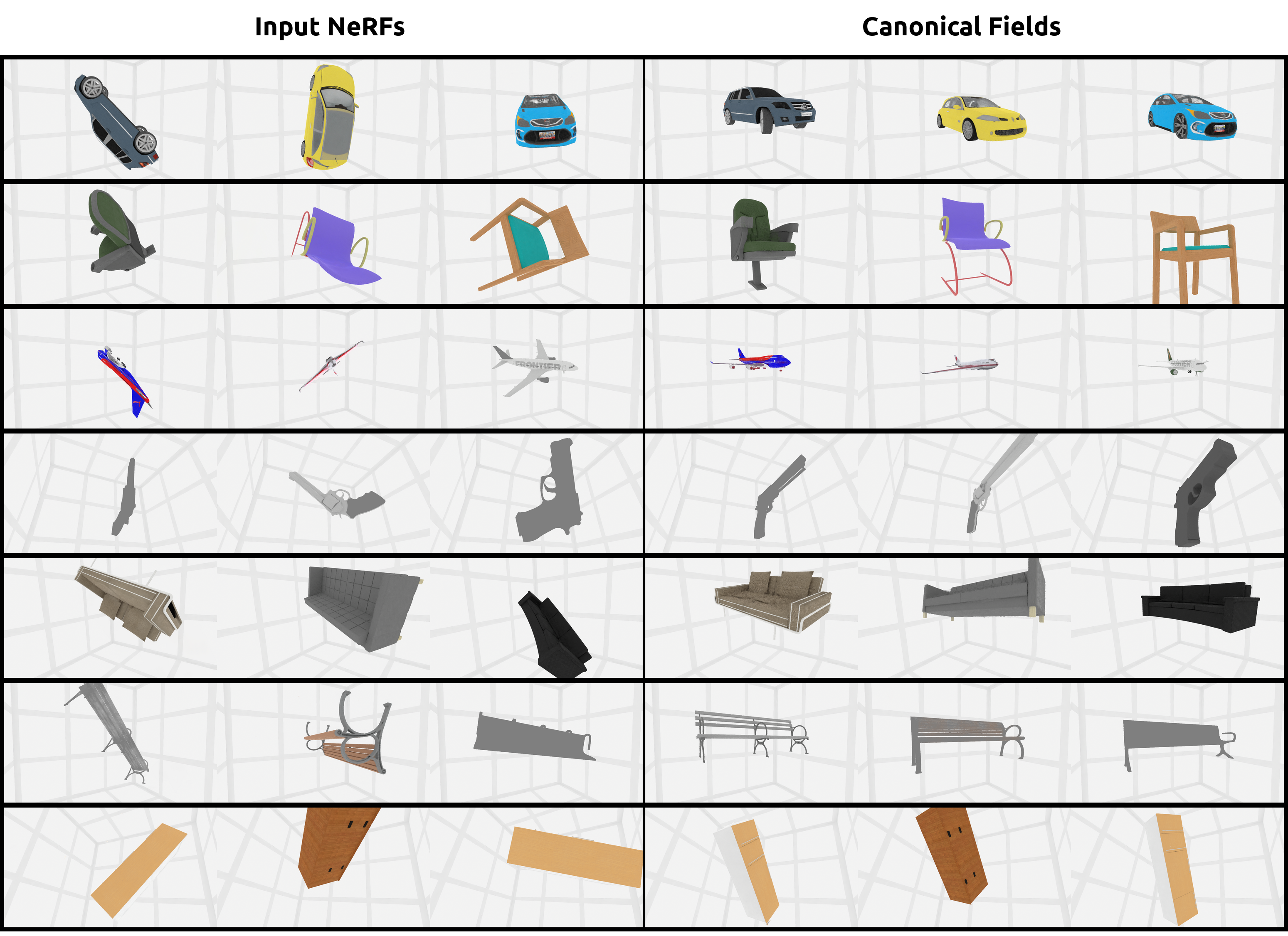}
    \caption{\textbf{\MethodName} qualitative canonicalization results for $7$ categories (see following pages for more results).}
    \label{fig:appendix_qualitative}
    \vspace{3mm}
\end{figure*}
\vspace{10pt}

\section{Preprocessing Density Field}
Before querying NeRF for occupancy, we do not have the extremities of the object. We query a uniform grid within the unit cube to obtain a $32^{3}$ grid density field to find the object bounds and the center of the object. The density values are normalized to range $[0, 1]$ (see page 3 in the main paper). We then perform an initial K-means clustering with $K=2$ to separate the foreground and background clusters. Note that this clustering is on the entire scene and is different from the clustering we perform while computing the losses. The clustering that we perform while computing the loss is within the object bounding box and not for the entire scene. The mean of query coordinate locations in the foreground cluster gives us the center of the object $\textbf{c}$. We then obtain the extremities from the foreground cluster and find its maximal diagonal length ($\bm{l}$). We re-sample the region within the bounding cube (at center $\bm{c}$ side length $\bm{l}$) and use it as input to \MethodName.

\begin{figure*}[h]
    \centering
    \includegraphics[trim={0.0cm 0cm 0cm 0},clip,width=1.0\textwidth]{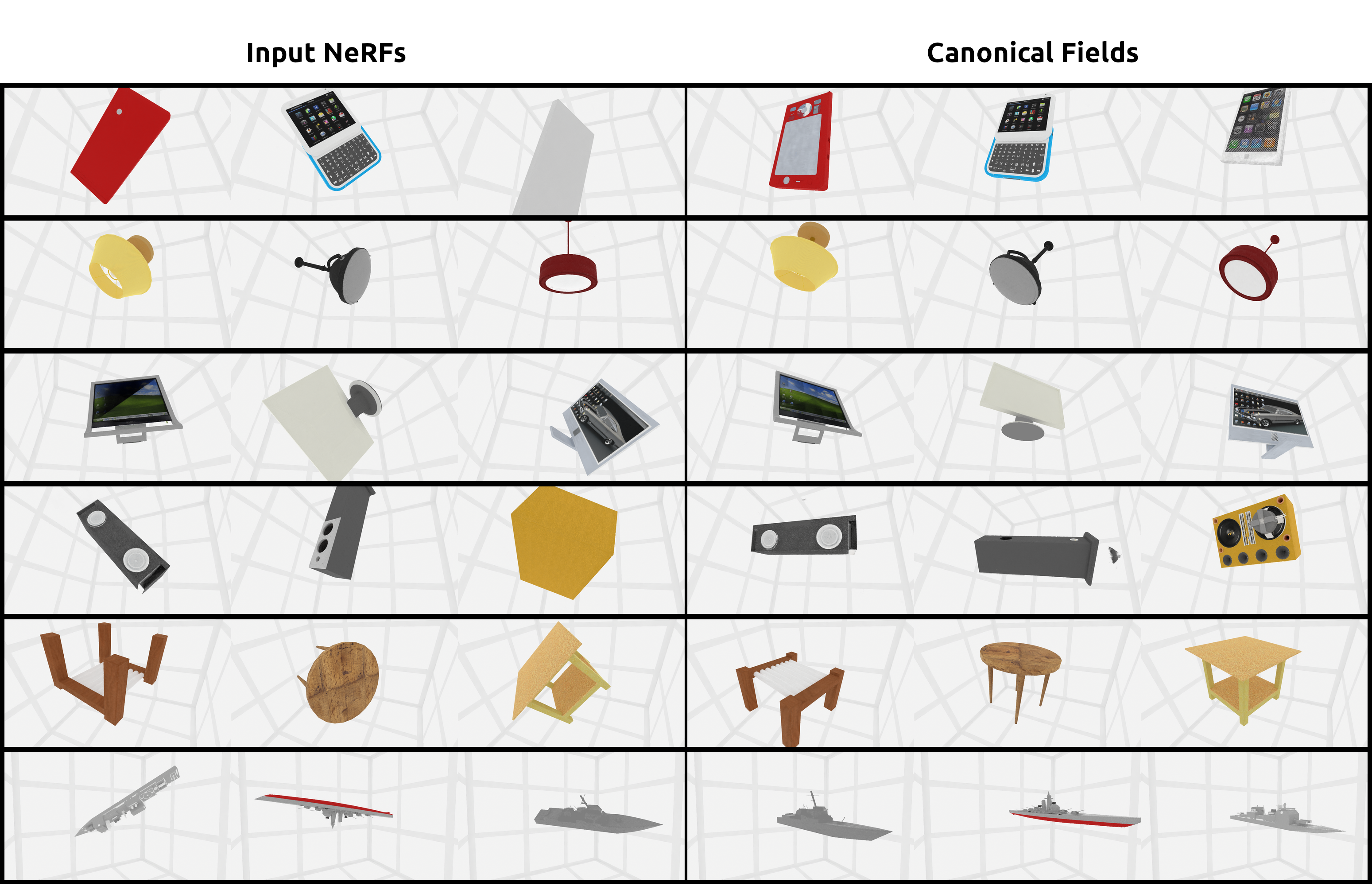}
    \caption{\textbf{\MethodName} qualitative canonicalization results for the remaining $6$ categories.}
    \label{fig:appendix_qualitative_2}
    \vspace{-3mm}
\end{figure*}

\section{Equivariance Properties of \MethodName}
\subsection{Averaging Equivariant Signals is Equivariant}



A tensor field of type $\ell$ ($\ell$-field) is a map $f:\mathbb{R}^3 \rightarrow \mathbb{R}^{2\ell + 1}$. We have an action of $\mathrm{SO}(3)$ on any $\ell$-field $f$ given by $(R.f)(x) := D^{\ell}(R)f(R^{-1}x)$ for any rotation $R \in \mathrm{SO}(3)$ and $x \in \mathbb{R}^3$ where $D^{\ell}(R) \in \mathrm{SO}(2\ell+1)$ is the Wigner matrix of type $\ell$. We say that transformation $F$ transforming tensor fields of type $p$ into tensor fields of type $q$ if it commutes with the action of $\mathrm{SO}(3)$, i.e. for all type $p$-field $f$ and rotation $R \in \mathrm{SO}(3)$ we have $F(R.f) = R.F(f)$. We will call such transformations $(p,q)$-equivariant transformations.
\begin{lemma}
    For any $r > 0$ the local average operator $\mathrm{mean}_r$ defined over $p$-fields by:
    \[
    \mathrm{mean}_r(f)(x) := \int_{B(x,r)} f(y) dy
    \]
    where $B(x,r) \subset \mathbb{R}^3$ is the open ball of radius $r$ centered at $x \in \mathbb{R}^3$ is a $(p,p)$-equivariant transform of fields.
\end{lemma}
\parahead{Proof} Let $x \in \mathbb{R}^3$ and $R \in \mathrm{SO}(3)$ we have:
\[
\begin{aligned}
\mathrm{mean}_r(R.f)(x)
&
=
\int_{B(x,r)} (R.f)(y) dy
\\
&
=
\int_{B(x,r)} D^p(R)f(R^{-1}y) dy
\\
&
\underset{u = R^{-1}y}{=}
D^p(R)\int_{R^{-1}B(x,r)} 
\hspace{-10mm}
f(u) du
\\
&
=
D^p(R)\int_{B(R^{-1}x,r)} 
\hspace{-10mm}
f(u) du
\\
&
=
(R.\mathrm{mean}_r(f))(x)
\end{aligned}
\]

\subsection{Locally Averaged Density Weighted Equivariant Vectors are Equivariant}
\begin{lemma}
Scaling an equivariant field with density is a (1, 1)-equivariant transformation.
\end{lemma}
\parahead{Proof} Let $\sigma$ be a type-$0$ density field and $f$ be a type-$1$ field at the corresponding location, weighing $f$ by the average of $\sigma$ is:
\[
\begin{aligned}
&\mathrm{mean}_r(\sigma)(x) \cdot f(x):= \left(\int_{B(x,r)} \sigma(y)dy\right) f(x) 
\\
& \mathrm{mean}_r(R \cdot \sigma)(x) \cdot (R\cdot f)(x) \\ 
& =\left(\int_{B(x, r)} (R\cdot \sigma)(y)dy\right) D^{1}(R) (f)(R^{-1}x) 
\\
& = \left( \int_{B(x, r)} (\sigma)(R^{-1}y)dy \right) D^{1}(R) (f)(R^{-1}x) 
\\
& = \left( I \cdot \int_{R^{-1}B(x, r)} (\sigma)(u)du \right) D^{1}(R) (f)(R^{-1}x) 
\\ 
& = D^{1}(R) \cdot \mathrm{mean}_r(\sigma)(x) \cdot (f)(R^{-1}x)
\end{aligned}
\]
Thus, $\mathrm{mean}_r(R\cdot \sigma)(x) \cdot f(x) = D^{1}(R) \cdot \mathrm{mean}_r(\sigma)(x) \cdot f(R^{-1}x)$ proving the result that scaling equivariant features by average density do not break the equivariance property.


\subsection{Gradient is Type-$1$ Equivariant}
\begin{lemma}
The gradient operator is a $(0, 1)$-equivariant transformation.
\end{lemma}

\parahead{Proof} Let $f$ be a $0$-type field, by the chain rule of differentiation, for any $x,h \in \mathbb{R}^3$ and $R \in \mathrm{SO}(3)$ we have
\[
\begin{aligned}
\langle \nabla (R.f), h \rangle
&
=
D_x (R.f)(h)
\\
&
= 
    D_x f \circ R^{-1}(h)
\\
&
=
D_{R^{-1}x} f
\circ
D_x R^{-1}(h)
\\
&
=
\langle \nabla_{R^{-1}x} f, R^{-1} h \rangle
\\
&
=
\langle R\nabla_{R^{-1}x} f,  h \rangle
\\
&
=
\langle R\nabla_{R^{-1}x} f,  h \rangle
\end{aligned}
\]
thus $\nabla (R.f) = R.\nabla f$ which concludes the proof.

\begin{table}[ht]
\centering
\scalebox{0.9}{
\begin{tabular}{c}
\toprule
\textbf{Equivariant Convolution Non-Linearities} \\
\midrule
\textbf{InverseSphericalHarmonicTransform}(sphere\_samples=$64$)\\
\textbf{BatchNorm}(momentum=$0.75$) \\
\textbf{ReLU} \\
\textbf{MLP}($F_{in}$, $F_{out}$)\\
\textbf{ForwardSphericalHarmonicTransform}(sphere\_samples=$64$)\\
\bottomrule
\end{tabular}
}
\caption{\textbf{Equivariant Convolution Non-Linearities} - We use the result in \cite{poulenard2021functional} and apply non-linearities after performing an inverse spherical harmonic transform to avoid breaking the equivariance of each layer. The \textit{sphere\_samples} is the sphere sampling resolution to perform the Spherical Harmonic Transform and $F_{in}, F_{out}$ are the input and output feature dimensions respectively.\label{tab:architecture}} 
\end{table}

\section{Additional Training Details}
To train \MethodName, we augment NeRF density fields with random rotations $R_{rand}$ and canonicalize them at each training iteration. We can easily do this by sampling the NeRF model at location $R_{rand}^{-1}x$ instead of $x$ in a differentiable manner using \cite{spatial_transformer}. Our method is built over equivariant layers and non-linearities from \cite{poulenard2021functional}. \cref{tab:architecture} shows the non-linearities and learning layers that are used after each equivariant convolution described in the main manuscript.


\section{Qualitative Results}
We illustrate qualitative results by rendering objects in the canonical frame for all the $13$ categories in \cref{fig:appendix_qualitative} and \cref{fig:appendix_qualitative_2}. Here, we fix a camera position and viewing direction in the canonical frame and render all objects from the same camera.

\end{document}